\documentclass[a4paper]{article}

\usepackage{INTERSPEECH2016}

\usepackage{graphicx}
\usepackage{amssymb,amsmath,bm,multicol}
\usepackage{textcomp}

\ninept

\title{NN-grams: Unifying neural network and n-gram language models \\ for 
	speech recognition}

\makeatletter
\def\name#1{\gdef\@name{#1\\}}
\makeatother 
\name{{\em Babak Damavandi, Shankar Kumar, Noam Shazeer, Antoine Bruguier}}
\address{Google Inc., 1600 Amphitheatre Parkway, Mountain View, CA 94043, USA \\
 {\small \tt \{babakd,shankarkumar,noam,tonybruguier\}@google.com}
}

\begin{document}

  \maketitle
  \begin{abstract}
    We present NN-grams, a novel, hybrid language model integrating n-grams and neural networks (NN) for speech recognition.  The model takes as input both word histories as well as n-gram counts. Thus, it combines the memorization capacity and scalability of an n-gram model with the generalization ability of neural networks. We report experiments where the model is trained on 26B words. NN-grams are efficient at run-time since they do not include an output soft-max layer. The model is trained using noise contrastive estimation (NCE), an approach that transforms the estimation problem of neural networks into one of binary classification between data samples and noise samples. We present results with noise samples derived from either an n-gram distribution or from speech recognition lattices. NN-grams outperforms an n-gram model on an Italian speech recognition dictation task.
  \end{abstract}

  \section{Introduction}
A \textit{language model} (LM) is a crucial component of natural language processing technologies such as speech recognition~\cite{jelinek97} and machine translation~\cite{brown90}. It helps discriminate between well-formed and ill-formed sentences in a language. Traditionally, n-gram LMs have formed the basis for most language modeling approaches. It has only been in the past few years that alternative approaches such as maximum-entropy models~\cite{chen09} and neural network models including feed-forward networks~\cite{bengio03,schwenk07,arisoy2012}, recurrent neural networks (RNNs)~\cite{mikolov11} and variants such as long short term memory (LSTM) networks~\cite{hochreiter97} have started outperforming n-gram models~\cite{jozefowicz2016}.

Neural network LMs have advantages over n-gram models. First, they provide better smoothing for rare and unknown words owing to their distributed word representations~\cite{jozefowicz2016}. Neural network Models such as LSTMs have the ability to remember long-distance context, an attribute that has eluded several language models in the past. Even with these potential advantages, LSTMs and other neural network models have not been used extensively for language modeling in speech recognition because they are more resource intensive at both training and run time when compared to n-gram models. This continues to be the case despite recent efforts at speeding up training and test times using techniques such as pipelined training and variance regularization~\cite{chen2015a}. LSTMs do not scale well to the large quantities of text training data typically used for estimating n-gram LMs. This is a substantial disadvantage during training because a larger LSTM which attains a better performance than a smaller LSTM is also slower to converge. At run time, an LSTM is expensive in terms of both memory and speed relative to an n-gram model. Specifically, the output soft-max layer is computationally expensive at both training and run time if the vocabulary size is in the order of millions of words, a common characteristic of current speech recognition systems (e.g. ~\cite{senior2015}).  Therefore, most neural network language modeling approaches for speech recognition have employed smaller vocabularies consisting of at most several thousands of words~\cite{schwenk07,mikolov11}. Speech recognizers for voice search and dictation typically operate on short utterances on which n-gram models perform fairly well. This has further limited the usefulness of LSTM LMs for these tasks.

In this paper, we investigate a flavor of neural network LMs that combines the strengths of a neural network in generalizing to novel contexts with the scalability and memorization ability of an n-gram model. Our main proposal is to train a neural network that is able to learn a mapping function given both the previous history of a given word as well as the n-gram counts, which are sufficient statistics for estimating an n-gram model. By providing n-gram counts as inputs, we expect this model to learn simultaneously a function of the counts as well as the word history and estimate the  probability of the current word given the history. Specifically, this model is a feed-forward neural network that takes as input the current word, $K$ previous words, and counts for the $N$ n-grams ending at the current word, where $N < K$. We call this model \textit{neural network-ngrams} (\textit{NN-grams}), to emphasize that it makes direct use of n-gram count statistics. While there have been earlier efforts at incorporating hashes of n-gram features as inputs to an RNN~\cite{mikolov11},  we are not aware of a neural network model that directly takes n-gram counts as inputs.

To reduce computation, we do not include an output soft-max layer in NN-grams. While the NN-grams' score can be interpreted as a log probability of the current word given the history, the absence of a soft-max layer means that these probabilities do not necessarily sum to one over the entire vocabulary.
 
\section{NN-grams}
\label{sec:model}
An LM is a probability distribution over the current word given the preceding words: $P(w_i | w_{i-1}, w_{i-2}, \ldots, w_{1})$. 
An n-gram LM makes the assumption that the current word depends only on the previous $N-1$ words: i.e.
\begin{equation*}
P(w_i | w_{i-1}, \ldots, w_{1}) = P(w_i|w_{i-1},\ldots,w_{i-(N-1)}).
\end{equation*} 
      \begin{figure}[t]
      	\centering
      	\includegraphics[height=1.8in]{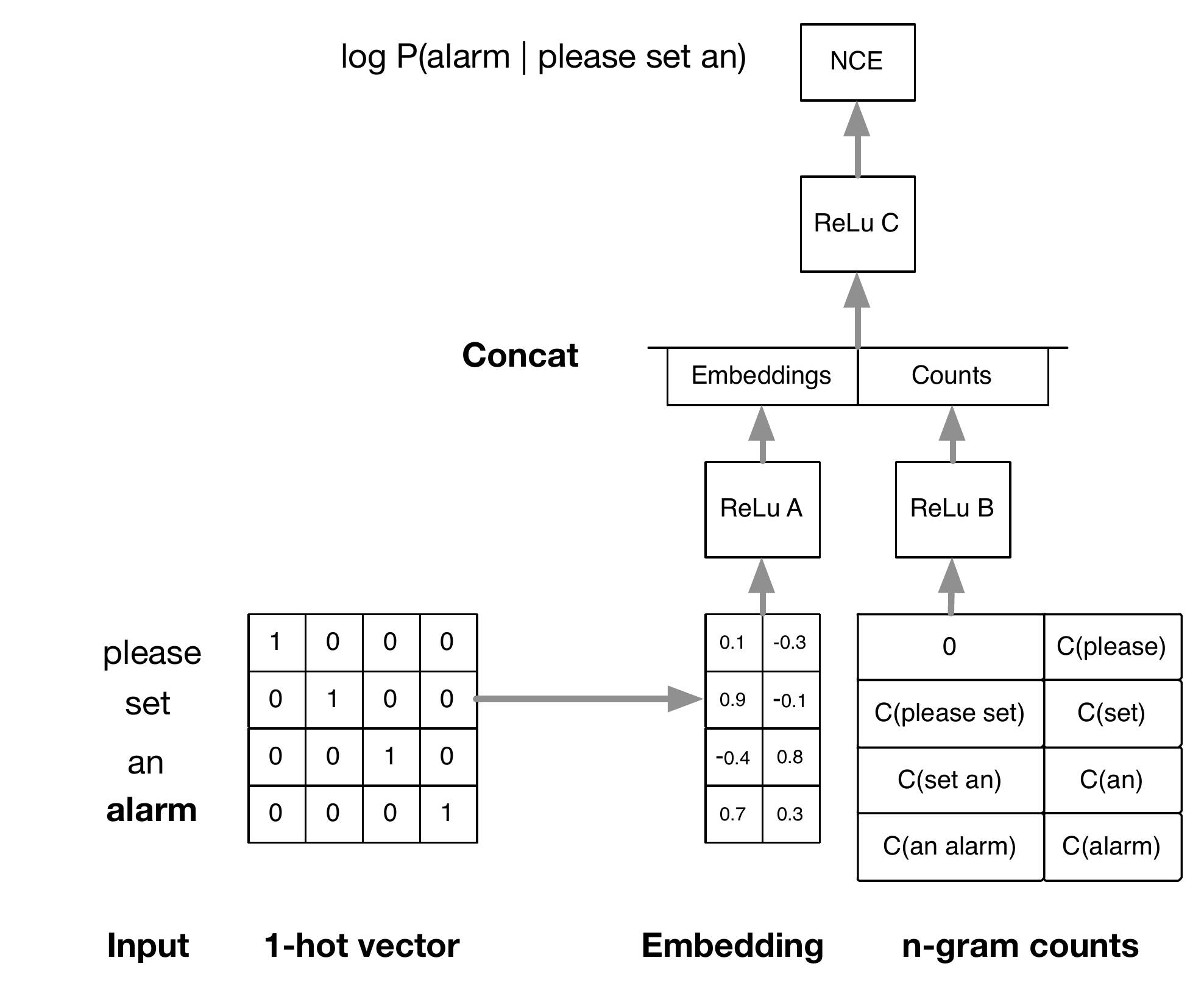}
      	\caption{{\it Architecture of the NN-grams model. The previous words are \textbf{``please'' ``set'' ``an''} and the current word is \textbf{alarm}. $N=2$, $K=3$ and the embedding dimension is 2.}}
      	\label{fig:nngrams}
      \end{figure}

The architecture of the NN-grams model is given in Figure~\ref{fig:nngrams}. The model takes as input the current word, $K$ preceding words and counts for the $N$ n-grams ending at the current word and estimates the log likelihood of the current word given the history:
 \begin{center}
 \scalebox{0.94}{
 $
    f_{\text{nng}}(w_i, \ldots, w_{1})
  = \text{log} P(w_i | w_{i-1},\ldots, w_{i-K}, \bf{c})
 $
 }
 \end{center}
where $\bf{c}$ is a vector of n-gram counts of length $(K+1)N$ such that $c_{1} = \text{Count}(w_i)$, $c_2=\text{Count}(w_i,w_{i-1})$,..., $C_{N}=\text{Count}(w_i,w_{i-1},\ldots,w_{i-(N-1)})$ are counts of n-grams ending at the current word, $c_{N+1}$ through $c_{2N}$ are counts of n-grams ending at the previous word,..., and $c_{KN+1}$ through $c_{(K+1)N}$ are counts of n-grams ending at word $w_{i-K}$ (See the count matrix in Figure~\ref{fig:nngrams} for an example), and the log probability is estimated by the neural network. Each of the words $w_{i}, \ldots, w_{i-K}$ is presented as a 1-hot vector to the network. The number of previous words, K, can be larger than N, the order of the n-gram counts. This enables the model to take into account longer context. Like other neural network LMs~\cite{bengio03}, the NN-grams model maps words into a high dimensional space and learns an \textit{embedding} for each word in this space while simultaneously also learning the network parameters. The word embeddings and the n-gram counts are passed through separate layers with  rectified linear unit (ReLu) activations~\cite{nair2010} and then concatenated. The result is passed through a third ReLu layer and provided as an input to NCE. The output of the NCE layer approximates the log probability of the current word given the history and the n-gram counts. Unlike other neural network language modeling approaches~\cite{bengio03,schwenk07,mnih12}, there is no explicit soft-max over the vocabulary.
     
  \subsection{Model Estimation}
  We train the neural network using noise contrastive estimation (NCE), a method for training unnormalized probabilistic models~\cite{gutmann10,mnih12,mnih13}. NCE transforms the estimation problem of the network into a classification problem where the goal is to differentiate between samples from the training data ($D=1$) and those from a pre-specified noise distribution ($D = 0$). For brevity, we abbreviate the current word, $w_i$ as $w$ and its history $w_{i-1} \ldots, w_{i-K}, \bf{c}$ as $h$. Our goal is to fit the neural network to the training data distribution $P_{\text{data}}(w|h)$. 

Suppose we have $f$ noise samples for each training data sample, the posterior probability that the sample $(h,w)$ arises from the training data is given by~\cite{mnih13}:
 \begin{equation*}
  P(D=1 | w, h) = \frac{ P_{\text{data}}(w | h)}{ P_{\text{data}}(w | h) + P_{\text{noise}}(w |h) f}.
  \end{equation*}
\noindent We estimate this probability by replacing the data distribution $\text{log} P_{\text{data}}(w|h)$ with that of the neural network $\text{NN}(w, h)$:
  \begin{align*}
   \text{logit} (D=1| w, h) & = & \text{log} \left( \frac{P(D=1|w,h)}{1-P(D=1|w,h)} \right)\\   
   & = & \text{log} P_{\text{data}} (w | h) - \text{log}(f) - \text{log} P_{\text{noise}} (w | h) \\ 
   & \approx & \text{NN} (w, h)  - \text{log}(f) - \text{log} {P}_{\text{noise}} (w | h).
  \end{align*}

\vspace{-0.1cm}
\begin{figure}[t]
      	\centering
      	\includegraphics[height=2.0in]{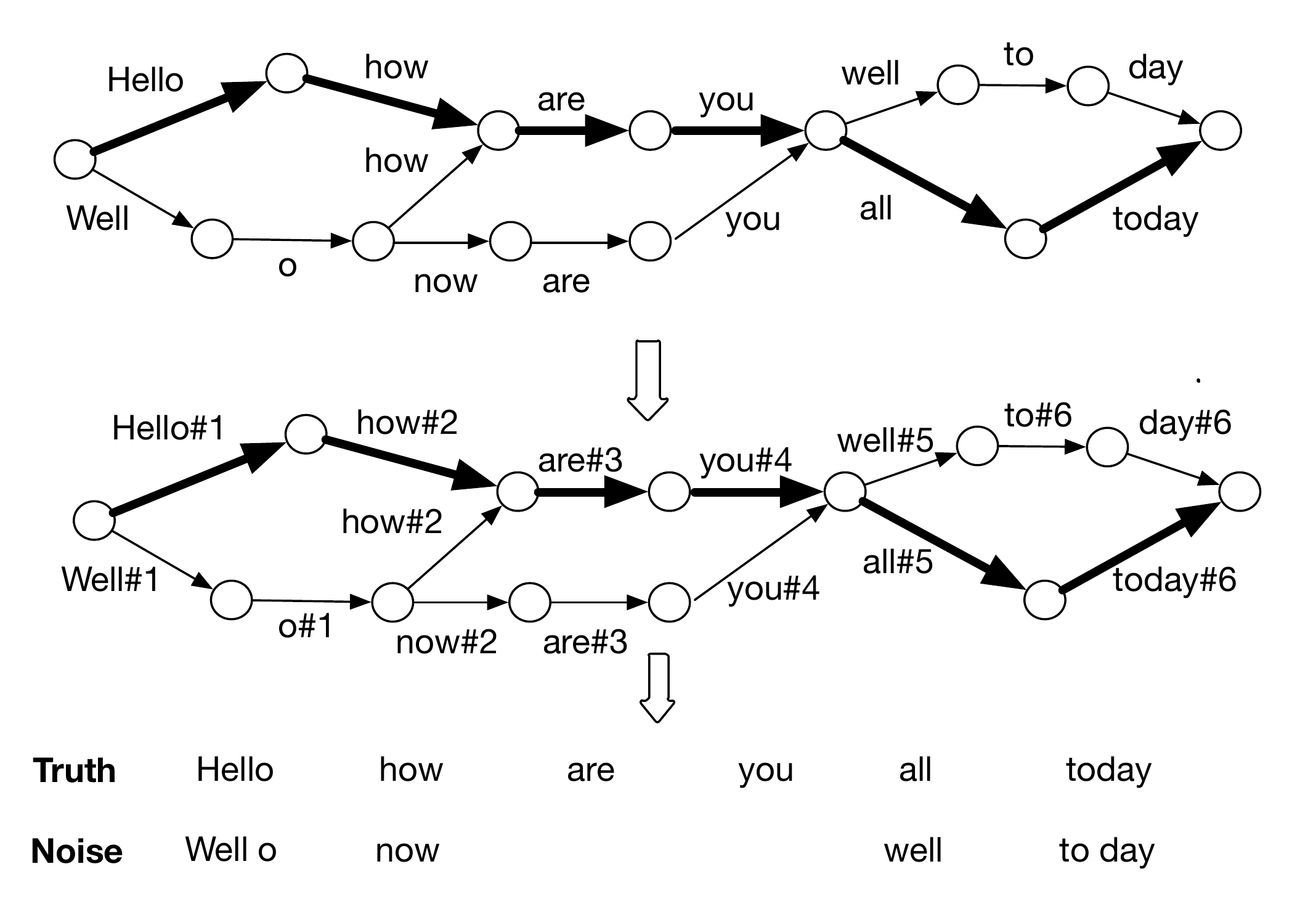}
      	\caption{{\it Extracting speech noise samples via lattice pinching. All hypotheses in a lattice (top panel) are aligned with respect to the 1-best hypothesis (shown in bold). For each lattice edge, the alignment relative to the 1-best hypothesis is determined (middle panel). The list of noise samples is then extracted for each position (bottom panel).}}
      	\label{fig:pinching}
\end{figure}
\vspace{-0.1in} 
 \subsection{Noise Distributions}
\label{sec:noise}
NCE training works best when the noise distribution is close to the data distribution. In this case, the training data samples are hard to distinguish from noise samples and the model is forced to learn about the structure of the data~\cite{gutmann10}. We experiment with two types of noise distributions. In the first type, we sample the noise word from the n-gram distribution over the words given the history. We will refer to this as \textit{text noise}. In the second type, we sample the noise word from the word level confusions generated from a speech recognition system. We will refer to this as \textit{speech noise}. Unlike text noise, speech noise consists of words which are acoustically confusable with the words in the training data. 

Ideally, these noise samples would be words which are transcribed incorrectly by the speech recognition system when compared with a human transcription. However, the quantity of human transcriptions is limited. Therefore, we run the recognizer on utterances where human transcriptions are not available and additionally, a 1-best recognition hypothesis with high confidence exists. The noise words are the alternatives to the 1-best recognition hypothesis. We align the $1$-best hypothesis to the paths in the recognition word lattice using lattice pinching~\cite{goel04} (Figure~\ref{fig:pinching}) and obtain a set of noise samples for each word in the 1-best hypothesis. Within each such set, the noise probability of a given word is its posterior probability. We exclude those words in the 1-best hypothesis which a) do not have confusions in the lattice e.g. \textit{are} and \textit{you} in Figure~\ref{fig:pinching}, and b) align to word sequences with more than more word e.g. \textit{Hello} aligns with \textit{well o} in Figure~\ref{fig:pinching}. 

 \subsection{Count Rescaling}
  One of the inputs to \emph{NN-grams} is a vector of $n$-gram counts. Since this count can have a large dynamic range from $0$ to several millions, we rescale the count to improve the convergence of neural network training using gradient descent~\cite{lecun98}.  If $C$ is the original count, the rescaled count is obtained as $C' = 0.1 \; \text{log}(C)$ if $C > 0$ and $-1$ if $C=0$.
   
 \section{Experiments}
\label{sec:exps}
 We evaluated the \emph{NN-gram} language model (LM) on Italian voice-search and dictation speech recognition tasks. Since the NN-gram model does not yield probability estimates that are guaranteed to be normalized, we do not report perplexities. Our test sets consisted of a voice-search (VS) set with 12,877 utterances (27.4 hours, 47,867 words) and a dictation (DTN) set with
 12,625 utterances (19.2 hours, 82,121 words). All utterances were anonymized. The acoustic models were trained using convolutional, LSTM, fully connected deep neural networks as described in~\cite{sainath2015}. All LMs were trained on anonymized and aggregated search queries and dictated texts. A 5-gram LM with Katz backoff was trained using a total of 26B words, and consisted of a total of 102M n-grams. The initial word lattice was generated using this 5-gram LM and a recognition vocabulary consisting of 3.9M words.
 
 The NN-grams model was trained on the same corpus as the 5-gram LM. Since the NN-grams model takes 6-gram counts as input, we additionally trained a 6-gram LM with Katz backoff to provide a fair baseline. Prior work~\cite{chelba2010,roark2013} has shown that when using pruning, n-gram models with Katz backoff outperform those with Kneser-Ney smoothing~\cite{kneser1995}. Hence, we used Katz backoff as the smoothing technique for all our n-gram language models. We limited the vocabulary size to 2M words for both models. Even though the NN-grams model has fewer parameters than the 6-gram LM (Table~\ref{tab:modelparams}), it requires the availability of n-gram counts at run time.

\begin{table}[h]
 \begin{center}
 \begin{tabular}{c|c|c} \hline
 LM & parameter type & \# of parameters \\ \hline
 6\-gram & n-grams & 9.6B  \\ 
 NN\-gram & NN parameters & 517M \\ \hline
 \end{tabular}
 \caption{Model Parameters of NN-grams and 6-gram LMs.}
 \label{tab:modelparams}
 \end{center}
 \end{table}
\vspace{-0.1in}

The word lattices generated in the initial recognition pass were rescored using either the 6-gram LM or the NN-grams model. In the case of the NN-grams LM, there is no exact algorithm for rescoring the lattice. We note that there have been approximate algorithms to rescore lattices using long-span neural network language models~\cite{sundermeyer2014, liu2014efficient}. However, we did not employ these lattice rescoring methods and instead, extracted and reranked the 150-best word hypotheses from the lattice.\footnote{If there were fewer than 150 hypotheses for an utterance, we extracted the maximum number of available hypotheses.} The score (log probability) of the either the 6-gram LM or the NN-grams model was interpolated with the log probability of the 5-gram LM using a fixed weight of 0.5. The 5-gram LM gave a Word Error Rate (WER) of 17.9\% on VS and 11.8\% on DTN.
 
We set the parameters $K$ and $N$ of the NN-grams model to 9 and 6 respectively. The model was trained until convergence with an AdaGrad optimizer~\cite{duchi2011} using a learning rate that was set to 0.01. We used a batch size of 200 in training. The dimensionality of the word embedding layer was 256. The ReLu layer that processed the embeddings (ReLu-A) had 1024 units while the Relu layer that processed the n-gram counts (ReLu-B) had 256 units.  Finally, the ReLu layer that processed the concatenation of embeddings and counts (ReLu-C) had 1024 units. For NCE, we generated one noise sample for each word in the training data using text noise.

\subsection{Comparison with n-gram LMs}
We first compared the performance of NN-grams with the 6-gram LM. The results are shown in Table~\ref{tab:ngramcompare}. When compared with the 6-gram LM, the NN-grams LM showed a better performance on the DTN task and an equivalent performance on the VS task. While additional gains might be potentially obtained by first rescoring the lattice with a 6-gram model followed by interpolation with NN-grams, such a system would be too slow to deploy in a speech recognition system with stringent latency requirements. Hence, we did not pursue such an interpolation.

\begin{table}[h]
\begin{center}	
\begin{tabular}{c|c|c}
\hline
& VS & DTN \\ \hline 
6-gram & 14.9 & 8.8 \\
NN-grams & 14.8 & 8.2 \\ \hline	
\end{tabular}
\caption{{\it WER Comparison of NN-grams with 6-gram LM on voice-search and dictation.}}
\label{tab:ngramcompare}
\end{center}
\end{table}
\vspace{-0.1in}

\subsection{NN-grams components}
NN-grams consist of two components: word embeddings and the n-gram counts. To determine which of these two components had a bigger impact on the overall performance of the NN-grams  model, we trained the model with either one of these inputs (Figure~\ref{fig:nngrams}). For both VS and DTN, n-gram counts were more important than word embeddings (Table~\ref{tab:nngramcomponents}). We expect this result considering that using n-gram counts typically improves the performance for short sentences, which is the case for both VS and DTN (Average number of words/sentence on VS and DTN is 3.7 and 6.5 respectively). For DTN, word embeddings contributed to an additional improvement in WER, that can be attributed to the longer sentence length in DTN.
\begin{table}[h]
\begin{center}
\begin{tabular}{c|c|c}
	\hline
	NN-grams components & VS & DTN \\ \hline 
	Word-embedding,n-gram counts & 14.8 & 8.2 \\
	Word Embedding & 15.3 & 8.8 \\
	n-gram Counts & 14.9 & 8.5 \\ \hline	
\end{tabular}
\caption{\label{tab:nngramcomponents} {\it Impact of NN-grams components on WER.}}
\end{center}
\end{table}
\vspace{-0.1in}

\begin{table*}[t]
{\footnotesize
 \begin{center}
\begin{tabular}{l|l|l} \hline
1 & Reference & Poi ha detto per le sagome dei marmi del bagno e che quando lui tornava dava le misure per fare fare i marmi \\
1 & n-gram  & poi ha detto per le sagome dei bei marmi del bagno \`{e} che quando lui torna da quale misura per fare fare i marmi \\
1 & NN-grams &  poi ha detto per le sagome dei bei marmi del bagno \`{e} che quando lui tornava dava le misure per fare fare i marmi \\ \hline
2 & Reference & \`{E} molto pi\`{u} forte rispetto alle altre classi tipo Audi, Mercedes \\
2 & n-gram & \`{e} molto più forte rispetto a Mercedes \\
2 & NN-grams & \`{e} molto più forte rispetto ad altre classi tv Audi Mercedes \\  \hline
3 & Reference & Amò mo mi puoi sposare C'abbiamo la casa c'abbiamo la chiesa e c'\`{e} la sposa \\
3 & n-gram & am\`{o} mi puoi sposare se abbiamo la casa che abbiamo la Chiesa Ecce la sposa \\
3 & NN-grams & am\`{o} mi puoi sposare c'abbiamo la casa c'abbiamo la Chiesa e c'\`{e} la sposa \\ \hline
 \end{tabular}
\caption{Examples of recognition hypotheses where the NN-grams LM outperforms the n-gram LM.}
 \label{tab:examples}
 \end{center}
}
\end{table*} 
\vspace{-0.1in}

\subsection{Type and Quantity of Noise Samples}
We next examined whether the type and number of noise samples influenced the performance of the NN-grams model (Table~\ref{tab:typenumnoise}). Since the speech noise samples can be obtained only from lattices, we restricted our training set in this experiment to only those utterances for which we were able to run the speech recognizer and generate word lattices. The training set consisted of 1.2B words from a subset of utterances derived from both voice search and dictation sources on which the 1-best recognition hypothesis had a high confidence. As a result, the WERs for these systems are worse than the system trained on 26B words (Table~\ref{tab:ngramcompare}). The speech and the text noise samples were generated using the procedure described in Sec~\ref{sec:noise}. The training data was annotated with n-gram counts derived from the 26B word corpus used in the earlier experiments.

For each type of noise, we report WER using 1, 5, 10 and 100 noise samples. For both types of noise, the best performance was seen at 100 samples per word. The text noise outperformed the speech noise on DTN but obtained an equivalent performance on VS. It is possible that text noise, that relies on an n-gram distribution, is more suited to the dictation task where long range context is useful. In contrast, the speech noise samples which are acoustically confusable alternatives do not always have long distance dependencies. Based on these results, we could expect additional gains using 100 noise samples in the original set up with 26B words (Table~\ref{tab:ngramcompare}).
\begin{table}
\begin{center}
\begin{tabular}{c|cc|cc}
\hline
\# of noise samples & \multicolumn{2}{c}{Text Noise} & \multicolumn{2}{c}{Speech Noise} \\ \hline
& VS & DTN & VS & DTN \\ \hline
1 &  19.2 & 11.6 & 21.7 & 14.5 \\ 
5 &  19.4 & 12.3 & 20.9 & 14.1 \\
10 & 20.1 & 12.6 & 20.4 & 13.9 \\
100 & 17.3 & 10.6 & 17.5 & 12.3 \\ \hline
\end{tabular}
\caption{Impact of the type and number of noise samples on WER.}
\label{tab:typenumnoise}
\end{center}
\end{table}

\subsection{Embeddings}
In the NN-grams model, each word is mapped to a real valued vector. These word embeddings are key to the generalization capabilities of a neural network LM.  We present examples of the top-5 nearest neighbors for two Italian words, \textit{Roma} and \textit{telefono} computed using the word embedding estimated in the NN-grams model (Table~\ref{tab:embeddings}). The nearest neighbors for \textit{Roma} are all cities in Italy while those for \textit{telefono} consist of terms related to communication and business. In general, these neighbors in the embedding space are related to the source word, thus emphasizing the semantic nature of the space.
 \begin{table}[h]
 \begin{center}
 	\begin{tabular}{cc|cc}
 	\multicolumn{2}{c}{Roma} & \multicolumn{2}{c}{telefono} \\ \hline
 	Word & ED & Word & ED  \\ \hline
         Bologna & 1.08 &  cellulare & 1.33 \\ 
 	 Milano & 1.09 & tel & 1.34 \\
 	 Firenze & 1.15 & contatti & 1.41 \\
 	 Torino & 1.16 & indirizzo & 1.47 \\
 	 Napoli & 1.17 & fax & 1.53 \\ \hline
 \end{tabular}
 		\caption{Top-5 nearest neighbors for two Italian source words: \textit{Roma} and \textit{telefono} computed using the NN-grams word embeddings. The Euclidean distance (ED) of each neighbor from the source word is also shown.}
 \label{tab:embeddings}
 \end{center}
\end{table}

\subsection{Examples}
We present example recognition hypotheses where the NN-grams LM substantially outperformed the 6-gram LM (Table~\ref{tab:examples}). In example 1, while the n-gram model prefers the common construction \textit{torna da (come back from)}, NN-grams is either recognizing a complex construction or prefers the tense agreement between \textit{tornava} and \textit{dava}. In example 2, the n-gram LM prefers dropping clauses while NN-grams does not. In example 3, the NN-grams model is possibly recognizing the repeating pattern in the sentence \textit{C'abbiamo la casa c'abbiamo la chiesa e c'\`{e} la sposa (we have a house, we have a church, we have a bride)} while the n-gram model looks independently at each of the 3 phrases, \textit{se abbiamo}, \textit{che abbiamo} and \textit{Ecce}, and misrecognizes all of them. This last example may be a scenario where the long 10-word window of NN-grams gives it a distinct advantage over the n-gram LM.

\section{Discussion}
\label{sec:discuss}
In this paper, we presented NN-grams, a novel neural network language modeling framework that builds upon the memorization capabilities and scalability of $n$-gram LMs while still allowing us to benefit from the generalization capabilities of neural networks.  Our model obtains a 7\% relative reduction in word error rate on an Italian dictation task. We showed that the strength of the NN-grams model comes primarily from the n-gram counts but both n-gram counts and word embeddings are important for long-form content such as dictation. We trained the model using NCE training with either text or speech noise distributions. While text noise is better for the dictation  task, both noise types perform similarly for voice-search. The biggest disadvantage of the speech noise approach is that it requires decoding of utterances. Future work will investigate strategies which can directly generate acoustically confusable noise samples from only text using strategies that have been investigated in the context of discriminative language modeling~\cite{kurata2012,jyothi2010}. These strategies generate noise samples at either the phonetic or sub-phonetic (e.g. Gaussian) levels. In conclusion, the NN-grams model is a promising neural network LM that is scalable to large training texts. By avoiding the output softmax layer, it has a substantially lower overhead at training and run time compared to current neural network approaches such as LSTMs. We expect that this model will spur newer hybrid architectures which will increase the adoption of neural network approaches to language modeling.

  \section{Acknowledgements}
  We would like to thank Kaisuke Nakajima, Xuedong Zhang, Francoise Beaufays, Chris Alberti and Rafal Jozefowicz for providing crucial support at various stages of this project.
  \newpage
  \eightpt
  \bibliographystyle{IEEEtran}
  \bibliography{paper}

\end{document}